\PassOptionsToPackage{dvipsnames}{xcolor}
\documentclass{article}
\usepackage[dvipsnames]{xcolor}
\usepackage{graphicx}
\usepackage{subcaption}
\usepackage[final]{corl_2026} 
\makeatletter
\renewcommand{\@conferencelocation}{Austin, Texas, USA}
\makeatother

\usepackage{amssymb}
\usepackage[utf8]{inputenc}
\usepackage{booktabs}    
\usepackage{multirow}    
\usepackage{amsmath}     
\usepackage{caption}     
\usepackage[capitalize]{cleveref}
\usepackage{enumitem}    
\usepackage{bm}          
\usepackage{layouts}          
\usepackage{siunitx}     

\crefname{section}{Sec.}{Secs.}
\crefname{figure}{Fig.}{Figs.}
\crefname{table}{Tab.}{Tabs.}
\crefname{equation}{Eq.}{Eqs.}

\usepackage[normalem]{ulem} 

\usepackage{xspace}
\makeatletter
\DeclareRobustCommand\onedot{\futurelet\@let@token\@onedot}
\def\@onedot{\ifx\@let@token.\else.\null\fi\xspace}

\makeatother

\newcommand{\mypara}[1]{\noindent\textbf{#1}\hspace{0.2em}}

\usepackage{titlesec}
\titlespacing*{\section}
  {0pt}    
  {1.4ex plus 0.2ex minus 0.1ex}  
  {0.7ex plus 0.1ex}              
\titlespacing*{\subsection}
  {0pt}    
  {0.7ex plus 0.2ex minus 0.1ex}  
  {0.35ex plus 0.1ex minus 0.1ex}  
\titlespacing*{\paragraph}
  {0pt}     
  {0.1ex plus 0.1ex minus 0.1ex}  
  {0.6em}     

\newcommand{\V}[1]{\bm{#1}}
\newcommand{\M}[1]{\mathbf{#1}}

\title{DeTac: Dual-branch Decoupling of Texture and Force Signals in Vision-Based Tactile Sensors}

\author{
  Zhirui Zhang$^{1,2}$ \quad 
  Yunkai Tang$^{1,2}$ \quad 
  Jiahang Li$^{1,2}$ \quad 
  Fan Fei$^{1,2}$ \quad 
  Jinnuo Zhang$^{3}$ \\
  \textbf{Kairan Ding$^{1,2}$ \quad 
  Jieji Ren$^{3,\dagger}$ \quad 
  Yandong Guo$^{2}$ \quad 
  Boxin Shi$^{1,\dagger}$} \\
  $^1$Peking University \quad 
  $^2$AI\textsuperscript{2} Robotics \quad 
  $^3$Shanghai Jiao Tong University \\
  \texttt{zrzhang25@stu.pku.edu.cn} \quad 
  \texttt{shiboxin@pku.edu.cn} 
}

\newcommand{\myPaperTitle}{DeTac: Dual-branch Decoupling of Texture and Force Signals in Vision-Based Tactile Sensors}

\def\maketitlesupplementary{
   \clearpage
   \begin{center}
      \centering
      \Large
      \textbf{\myPaperTitle} \\
      \vspace{0.5em}
      Appendix \\
      \vspace{1.0em}
   \end{center}
}

\usepackage[resetlabels]{multibib}
\newcites{supp}{Supplementary References}

\graphicspath{{DeTac/}}
\hypersetup{pdftitle={DeTac: Dual-branch Decoupling of Texture and Force Signals in Vision-Based Tactile Sensors},
 pdfauthor={Zhirui Zhang, Yunkai Tang, Jiahang Li, Fan Fei, Jinnuo Zhang, Kairan Ding, Jieji Ren, Yandong Guo, Boxin Shi}}
\begin{document}
\maketitle

\begingroup
\renewcommand{\thefootnote}{}
\footnotetext{$^\dagger$Corresponding authors. \textsuperscript{1}Zhirui Zhang, Yunkai Tang, Jiahang Li, Fan Fei, Kairan Ding, and Boxin Shi are with State Key Laboratory of Multimedia Information Processing, National Engineering Research Center of Visual Technology, and PKU-AI\textsuperscript{2} Robotics Joint Lab of Embodied AI, School of Computer Science, Peking University. \textsuperscript{3}Jinnuo Zhang and Jieji Ren are with the State Key Laboratory of Mechanical System and Vibration, School of Mechanical Engineering, Shanghai Jiao Tong University, Shanghai, and also with Shanghai Key Laboratory of Intelligent Robotics, Shanghai Jiao Tong University, Shanghai.}
\endgroup


\vspace{-10pt}
\begin{abstract}
Vision-based tactile sensors enable fine-grained perception for dexterous robotic manipulation, but current marker-based designs tightly couple surface texture and contact force signals within a single observation. 
Existing data-driven decoupling methods rely on serial mask-based inpainting, which suffers from mask-sizing dilemmas caused by physical indentation effects, error retention, and lack of real paired data for supervision. 
To address these challenges, we propose DeTac, a novel dual-branch generative framework that concurrently extracts clean shading and marker images from a single mixed tactile image. 
DeTac leverages full-image synthesis to naturally bypass mask dependency and introduces a cross-branch synergistic optimization that enables mutual supervision, effectively preventing error retention. 
Furthermore, we construct a strictly aligned, real-world paired tactile dataset using custom hardware to provide high-quality supervision. 
Extensive experiments demonstrate that DeTac achieves state-of-the-art decoupling accuracy, significantly facilitating downstream surface reconstruction and force estimation tasks while exhibiting cross-sensor generalization.
\end{abstract}



\section{Introduction}
\label{sec:intro}

Vision-based tactile sensors \citep{gelsight_previous_work} transform local contact signals into visible elastomer deformations under structured light, providing robots with fine-grained tactile perception crucial for dexterous manipulation \citep{VTLA, tactile-vla, reactive25, unitac2pose25, vt25, learning24, 3D-ViTac24}.
While markerless sensors capture high-resolution shading images to reconstruct surface geometric textures \citep{gelsight, gelslim}, their lack of spatial references hinders direct contact force estimation \citep{measurement}. 
To address this, marker-based sensors embed physical markers into the elastomer \citep{measurement, gelslim2}, tracking their displacements to calculate force \citep{gelsight_wedge, gelslim3}. 
However, this design tightly \textit{couples} texture and force signals within a single observation, as the markers inherently obstruct the surface shading that encodes geometric textures \citep{magicskin, muxgel}. 
Consequently, accurately \textit{decoupling} these signals to reliably recover both surface texture and contact force has become a critical challenge.


\begin{figure}
    \centering
    \includegraphics[width=\linewidth]{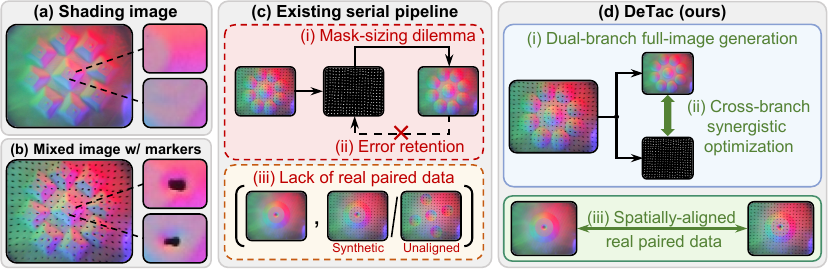}
    \caption{(a) Markerless sensors capture shading images to reconstruct surface but lack spatial references for force estimation. 
    (b) Marker-based sensors enable force sensing via embedded markers, but introduce coupled signals and indentation artifacts. 
    (c) Serial pipelines suffer from mask-size dilemmas, error retention, and lack of real paired training data. (d) DeTac uses a dual-branch framework with cross-branch optimization, trained on a self-collected aligned paired real dataset.}
    \label{fig:teaser}
\end{figure}

While early attempts at signal decoupling relied on complex hardware for mode switching \citep{UVtac22}, the rise of generative models \citep{DDPM, diffusion21, stable_diffusion} has enabled data-driven decoupling directly from a single mixed image. 
Existing data-driven methods \citep{markerless24, MVTS} typically employ a \textit{serial} pipeline: they first estimate a mask to locate the markers, then apply image inpainting \citep{telea, pix2pix} to infer from surrounding regions the occluded shading region as if the markers did not exist.
However, we observe that the physical embedding of markers, whether via sprayed ink \citep{gelsight} or 3D-printed pins \citep{tactip-family}, not only occludes the regions beneath the markers but also induces \textit{indentation artifacts} (\cref{fig:teaser}(a) vs. (b)): It alters the surface normals of the surrounding elastomer and therefore disrupts the perceived geometry texture at these regions.
This phenomenon results in three major limitations in current pipelines (\cref{fig:teaser}(c)). 
First, mask-based inpainting is confused by \textbf{mask-sizing dilemma}: Failing to cover the altered-normal regions (which vary with force and position) corrupts the predicted shading, while oversized masks discard critical texture information \citep{inpainting_mask, context_encoders}. 
Second, serial prediction suffers from \textbf{error retention}: Initial errors in marker localization propagate to the shading restoration step without any feedback mechanism to correct them.
Third, training is bottlenecked by the \textbf{lack of real paired data}: As strict alignment is difficult given real sensor fabrication variances, prior works rely on either synthetic marker superimposition that ignores indentation artifacts \citep{markerless24} or cycle-consistency using unaligned data \citep{bidirectional, unsupervised}, both yielding suboptimal real-world performance.

To overcome these limitations, we propose \underline{DeTac}, a \textit{dual-branch} framework that \underline{De}couples mixed \underline{Tac}tile observations to a marker-free shading image and a marker image (\cref{fig:teaser}(d)).
Unlike mask-based inpainting, DeTac treats shading restoration and marker extraction as \textbf{concurrent full-image generation} tasks, adaptively modifying regions without relying on predefined masks, thus naturally bypassing the mask-sizing dilemma.
We further introduce a \textbf{cross-branch synergistic optimization} that connects the two branches by enforcing spatial consistency between the predicted marker image and the restoration-induced shading changes, enabling mutual supervision that refines both predictions. 
Finally, to provide high-quality supervision, we construct a \textbf{real-world and strictly aligned paired dataset} using a specialized mechanical structure that captures exactly matched data on the same sensor before and after marker fabrication. 
In summary, our main contributions are:
\begin{itemize}[itemsep=0pt, topsep=0pt, leftmargin=1em]
\item We propose DeTac, a novel dual-branch generative framework that uses full-image synthesis to decouple mixed tactile signals, jointly predicting clean shading and marker images.
\item We introduce a cross-branch synergistic optimization that enables mutual supervision between shading restoration and marker extraction, effectively preventing error retention.
\item We construct a strictly aligned, real-world paired tactile dataset using a specialized mechanical structure, providing high-quality supervision for data-driven methods.
\end{itemize}
Extensive experiments demonstrate that DeTac achieves state-of-the-art accuracy in both shading restoration (with an over 7 dB improvement in PSNR over TacDiff~\citep{markerless24}) and marker extraction. 
Furthermore, the precisely decoupled signals significantly facilitate downstream tasks, including surface reconstruction and force estimation, while exhibiting generalization capabilities across tactile sensors.


\section{Related Work}
\label{sec:related_work}

\subsection{Vision-based Tactile Sensors}
Vision-based tactile sensors~\citep{gelsight, gelslim, digit, tactip-family} provide detailed tactile feedback by transforming local contact information into visible deformations of a soft, transparent elastomer~\citep{elastomeric}.
Early markerless designs~\citep{gelsight_previous_work, omnitact, soft-bubble} employ structured light to capture high-frequency geometric details of the contact interface, thereby preserving rich textural information.
Marker-based sensors introduce dedicated fiducial markers to complement these designs with force estimation capabilities.
GelSight~\citep{gelsight, improved, measurement}, GelSlim~\citep{gelslim2, gelslim3}, and DIGIT~\citep{digit} fabricate markers by spraying or transferring 2D black ink arrays onto the elastomer surface, while the TacTip family \citep{tactip, taccylinde, tactip-m2, tactip-family} utilizes dual-material 3D printing technology to monolithically manufacture biomimetic internal pins.
Due to the indentation artifacts introduced in \cref{sec:intro}, the resulting sensor output is a complex mixture of shading and marker, where texture and force signals are inherently conflated.
UVTac~\citep{UVtac22} addresses this by physically separating the two signals at the hardware output level, but introduces complex hardware overhead and limits the sensor to capturing only one modality at a time. 
TacDiff~\citep{markerless24} attempts to decouple the two signals computationally, but is constrained by serial pipeline architectures and the scarcity of paired ground-truth (GT) data.
These limitations motivate our dual-branch framework, which jointly recovers both modalities via cross-branch supervision, trained on a highly aligned paired dataset.

\subsection{Conditional Image-to-image Generation}
Conditional image-to-image generation synthesizes a target image guided by dense spatial or structural cues from a source image.
Early work tackled this through generative adversarial network (GAN)~\citep{GAN}-based frameworks~\citep{pix2pix, CycleGAN17}, which framed the task as a direct domain mapping problem.
However, due to the training instabilities inherent to GANs~\citep{GAN}, recent approaches have largely shifted toward diffusion models~\citep{DDPM}.
Among these, masked inpainting methods regenerate only specific regions by conditioning on the unmasked surrounding context, either by concatenating the masked image and mask as additional input channels~\citep{stableinpainting23} or by employing a separate feature extractor~\citep{BrushNet24}.
However, these methods require the undesired and desired signal components to be strictly separable at pixel level; otherwise, information loss or quality degradation is inevitable. 
Full-image generation methods avoid this limitation by generating the complete output directly from the source condition.
ControlNet~\citep{ControlNet23}, T2I-Adapter~\citep{T2IAdapter24}, and Uni-ControlNet~\citep{Unicontrolnet23} inject features extracted from the condition via dedicated encoder modules into a frozen diffusion backbone.
Palette~\citep{Palette22} and InstructPix2Pix~\citep{instructpix2pix23} instead train end-to-end, concatenating the source image with the noisy latent.
SDEdit~\citep{SDEdit} and Plug-and-Play Diffusion Features~\citep{plugandplay23} manipulate a pre-trained model's sampling process without training, while BBDM~\citep{BBDM23} reformulates the diffusion trajectory as a stochastic bridge between domains. 
We adopt full-image generation because the indentation artifacts cannot be separated at pixel level, making mask-based inpainting fundamentally unsuitable.

\begin{figure}
    \centering
    \includegraphics[width=\linewidth]{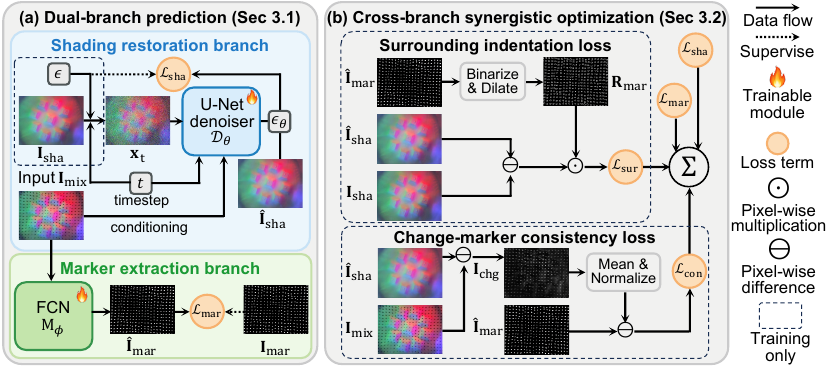}
    \caption{Overall pipeline of DeTac. 
    (a) DeTac takes as input a mixed tactile image and consists of two parallel branches (\cref{subsec:arch}): a conditional diffusion-based shading restoration branch and a lightweight marker extraction branch.
    (b) The cross-branch synergistic optimization (\cref{subsec:optimization}) exploits marker-aware spatial structures for mutual supervision using surrounding indentation loss for shading restoration and change-marker consistency loss for marker extraction.}
    \label{fig:pipeline}
\end{figure}

\section{Method}
\label{sec:method}

\cref{fig:pipeline} illustrates the overall pipeline of DeTac, 
whose (a) dual-branch architecture and (b) joint optimization are introduced in \cref{subsec:arch} and \cref{subsec:optimization}, respectively.
The hardware-aligned data collection process of the real-world paired dataset, on which DeTac is trained, is detailed in \cref{subsec: dataset} and \cref{fig:hardware}.

\subsection{Concurrent Dual-Branch Prediction}
\label{subsec:arch}

Given a mixture image $\M{I}_{\rm mix} \in \mathbb{R}^{H\times W\times 3}$ captured inside a marker-based visual-tactile sensor (the image is RGB as the structured-light illumination is color-coded), our goal is to decouple the coupled tactile signals into a marker-free shading image $\hat{\M{I}}_{\rm sha} \in \mathbb{R}^{H\times W\times 3}$ for surface geometry reconstruction and a marker image $\hat{\M{I}}_{\rm mar} \in [0,1]^{H\times W}$ for marker tracking and force estimation.
$\hat{\M{I}}_{\rm mar}$ is a soft mask image to account for defocus blur and ambiguous marker boundaries near the image periphery.

Previous methods rely on serial pipelines \citep{markerless24}, where a marker mask is estimated and used for mask-based inpainting to obtain the shading image, therefore suffering from the aforementioned mask-sizing dilemma and error retention.
To address these limitations, DeTac adopts a concurrent dual-branch architecture: 
The \textit{shading restoration branch} performs conditional full-image generation to recover the shading image, while the \textit{marker extraction branch} directly predicts the marker image.


\mypara{Shading Restoration Branch.}
The shading restoration branch recovers the marker-free shading image $\hat{\M{I}}_{\rm sha}$ by restoring marker-occluded regions and suppressing marker-induced indentation artifacts in the mixed input $\M{I}_{\rm mix}$.
We formulate this task as conditional full-image generation with a denoising diffusion model \citep{DDPM}, allowing the model to exploit all available information in $\M{I}_{\rm mix}$ rather than relying on a predefined inpainting mask and discarding the marker-affected pixels as in mask-guided inpainting methods \citep{markerless24}.
Specifically, our U-Net denoiser $D_{\theta}$ \citep{unet} takes the noisy target $\V{x}_t$, the timestep $t$, and the mixed image $\M{I}_{\rm mix}$ as input at each discrete diffusion timestep $t\in\{1,\ldots,T\}$:
\begin{equation}
\V{\epsilon}_{\theta}
=
D_{\theta}(\V{x}_t,t,\M{I}_{\rm mix}).
\end{equation}
We use a predefined variance schedule $\{\beta_t\}_{t=1}^{T}$ with $0<\beta_t<1$, and define $\alpha_t=1-\beta_t$ and $\bar{\alpha}_t=\prod_{s=1}^{t}\alpha_s$. 
During training, for a sampled timestep $t$, the noisy target $\V{x}_t$ is obtained as
\begin{equation}
\label{eq:noisy_sample}
\V{x}_t
=
\sqrt{\bar{\alpha}_t}\V{x}_0
+
\sqrt{1-\bar{\alpha}_t}\V{\epsilon},
\end{equation}
where $\V{x}_0=\M{I}_{\rm sha}$ denotes the GT noise-free target shading image, and $\V{\epsilon} \sim \mathcal{N}(\V{0}, \M{I})$ is Gaussian noise. 
The denoiser is optimized using the standard noise-prediction objective:
\begin{equation}
\label{eq:diffusion_loss}
\mathcal{L}_{\rm sha}
=
\mathbb{E}_{(\M{I}_{\rm mix},\M{I}_{\rm sha}),\V{\epsilon},t}
\left[
\left\|
\V{\epsilon}
-
D_{\theta}
\left(
\sqrt{\bar{\alpha}_t}\M{I}_{\rm sha}
+
\sqrt{1-\bar{\alpha}_t}\V{\epsilon},
t,
\M{I}_{\rm mix}
\right)
\right\|_2^2
\right].
\end{equation}
To compute auxiliary losses (see \cref{subsec:optimization}) during training, the clean image is estimated from $\V{\epsilon}_{\theta}$ as
\begin{equation}
\label{eq:x0_estimation}
\hat{\V{x}}_0
=
(\V{x}_t-\sqrt{1-\bar{\alpha}_t}\V{\epsilon}_{\theta})/\sqrt{\bar{\alpha}_t},
\quad
\hat{\M{I}}_{\rm sha}=\hat{\V{x}}_0.
\end{equation}
During inference, the model starts from pure Gaussian noise and iteratively samples $\V{x}_{t-1}$ from $\V{x}_t$:
\begin{equation}
\label{eq:diffusion_inference}
\V{x}_{t-1}
=
\left(
\V{x}_t
-
(1-\alpha_t)\V{\epsilon}_{\theta}/\sqrt{1-\bar{\alpha}_t}
\right)/\sqrt{\alpha_t}
+
\sigma_t\V{z}, \qquad t=T,\ldots,1,
\end{equation}
where $\V{z} \sim \mathcal{N}(\V{0},\M{I})$.
Following DDPM \citep{DDPM}, we set $\sigma_t^2=(1-\bar{\alpha}_{t-1})\beta_t/(1-\bar{\alpha}_t)$ with $\bar{\alpha}_0=1$. 
The restored shading image is given by the final denoised sample $\hat{\M{I}}_{\rm sha}=\hat{\V{x}}_0$.

\mypara{Marker Extraction Branch.}
The marker extraction branch isolates marker regions from the same mixed input $\M{I}_{\rm mix}$. 
The extracted marker image encodes the spatial distribution of markers under contact-induced deformation, from which marker-level quantities such as centroid positions can be obtained through post-processing. 
These quantities can then be used by downstream force-estimation algorithms \citep{measurement}, for example, by comparing displaced marker positions with their reference positions under the no-contact state.
Specifically, we employ a lightweight fully convolutional network $M_{\phi}$ \citep{FCN} to predict a soft marker image in a deterministic forward pass: $\hat{\M{I}}_{\rm mar}=M_{\phi}(\M{I}_{\rm mix})$.

During training, given the pseudo GT marker image $\M{I}_{\rm mar}$, whose acquisition will be detailed in \cref{subsec: dataset}, this branch is supervised using a pixel-wise Binary Cross-Entropy loss:
\begin{equation}
\label{eq:marker_loss}
\mathcal{L}_{\rm mar}
=
\mathrm{BCE}(\hat{\M{I}}_{\rm mar},\M{I}_{\rm mar}).
\end{equation}

\subsection{Cross-Branch Synergistic Optimization}
\label{subsec:optimization}

Although the shading branch and the marker branch predict different targets, both are governed by the same marker distribution and its influence on the tactile image. 
We therefore introduce a cross-branch synergistic optimization that explicitly connects the two branches through marker-aware spatial structures. 
This optimization consists of two complementary objectives: a \textit{surrounding indentation loss} that emphasizes restoration around marker neighborhoods, and a \textit{change-marker consistency loss} that aligns marker prediction with restoration-induced appearance changes.

\begin{figure}[!t]
    \centering
    \includegraphics[width=\textwidth]{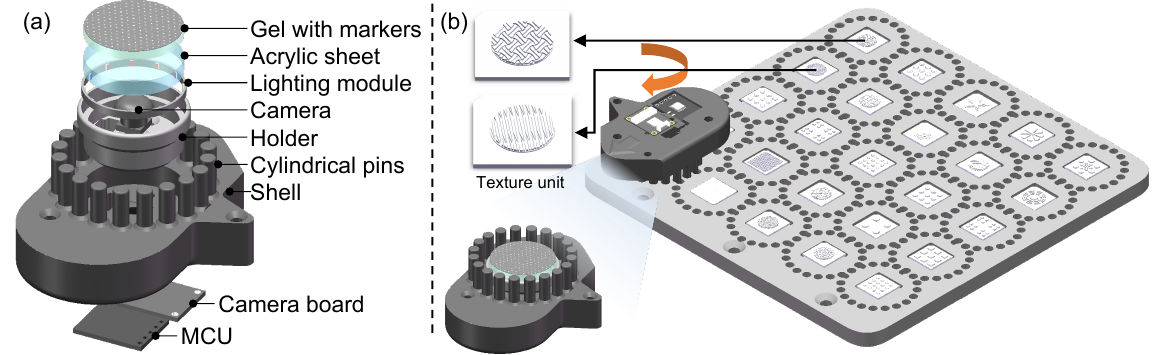}
    \caption{Hardware design and data acquisition setup. 
    (a) Exploded view of the MC-Tac \citep{mc_tac} sensor equipped with an outer shell featuring 20 cylindrical alignment pins arranged in a ring for repeatable positioning.
    (b) The data acquisition system features a collection pad with a 5 $\times$ 5 array of interchangeable texture units, each surrounded by 20 holes that mate with the sensor's alignment pins. Rotating the sensor and fully inserting the pins into different indexed holes under consistent force enable repeatable sampling of 20 contact poses per object before and after marker fabrication.}
    \label{fig:hardware}
\end{figure}

\mypara{Surrounding Indentation Loss.}
In tactile images, marker-induced artifacts are not limited to marker-covered pixels. 
Due to local indentation introduced during marker fabrication, the neighborhood around each marker is often difficult to restore faithfully. 
To emphasize these critical regions, we construct an annular neighborhood around the predicted marker regions and apply focused supervision there.
Specifically, let $\M{B}_{\rm mar}=\mathbb{I}(\hat{\M{I}}_{\rm mar}>\tau)$ denote the binarized marker prediction with threshold $\tau$. 
The annular neighborhood around the predicted marker regions is defined as
\begin{equation}
\M{R}_{\rm mar} = \mathrm{Dilate}(\M{B}_{\rm mar}, r) \setminus \M{B}_{\rm mar},
\label{eq:ring_region}
\end{equation}
where $\mathrm{Dilate}(\cdot,r)$ denotes morphological dilation with radius $r$, and $\setminus$ denotes set difference. 
Based on this ring-shaped region, we define the surrounding indentation loss as
\begin{equation}
\mathcal{L}_{\rm sur}
=
\|
(\hat{\M{I}}_{\rm sha} - \M{I}_{\rm sha})\odot \M{R}_{\rm mar}
\|_1
/ \left(\left\| \M{R}_{\rm mar}  \right\|_1 +\delta\right), 
\label{eq:sur_loss}
\end{equation}
where $\odot$ denotes element-wise multiplication, $\M{R}_{\rm mar}$ is broadcast to RGB channels, and $\delta$ is a small constant. 
This objective concentrates restoration supervision on marker neighborhoods, encouraging faithful reconstruction in the regions most affected by marker-induced indentation.


\mypara{Change-Marker Consistency Loss.}
The pseudo GT marker images generated by SAM \citep{sam2} with manual point annotations (see \cref{subsec: dataset}) may be inaccurate around marker boundaries, making $\mathcal{L}_{\rm mar}$ alone insufficient for marker supervision. 
To provide complementary guidance, we introduce a change-marker consistency loss $\mathcal{L}_{\rm con}$ that exploits the correlation between marker locations and restoration-induced appearance changes to provide mutual supervision.

We first compute a change map $\M{I}_{\rm chg}\in[0,1]^{H\times W}$ from the difference between the restored shading image and the mixed image. 
Specifically, we average the RGB difference across channels, sharpen the response with an exponent $\gamma=2$, and normalize it spatially:
\begin{equation}
\bar{\M{I}}_{\rm chg}
=
\operatorname{mean}_{\rm c}
(
|
\hat{\M{I}}_{\rm sha} - \M{I}_{\rm mix}
|
), 
\qquad
\M{I}_{\rm chg}
=
\bar{\M{I}}_{\rm chg}^\gamma/(\operatorname{max}_{\rm s}(\bar{\M{I}}_{\rm chg}^\gamma)+\epsilon),
\label{eq:change_map}
\end{equation}
where $\operatorname{mean}_{\rm c}$ denotes channel-wise averaging, $\max_{\rm s}$ denotes the spatial maximum, and $\epsilon$ is a small constant. 
The change-marker consistency loss is then defined as
\begin{equation}
\mathcal{L}_{\rm con}
=
\|
\hat{\M{I}}_{\rm mar} - \M{I}_{\rm chg}
\|_1.
\label{eq:con_loss}
\end{equation}
Through this synergistic optimization, the two branches are connected through shared marker-aware spatial cues, improving both shading restoration fidelity and marker extraction accuracy.

\mypara{Overall Objective.}
Together with the diffusion loss $\mathcal{L}_{\rm sha}$ and the marker extraction loss $\mathcal{L}_{\rm mar}$, the two auxiliary losses form the final training objective:
\begin{equation}
\mathcal{L}
=
\mathcal{L}_{\rm sha}
+
\lambda_{\rm mar} \mathcal{L}_{\rm mar}
+
\lambda_{\rm sur} \mathcal{L}_{\rm sur}
+
\lambda_{\rm con} \mathcal{L}_{\rm con},
\label{eq:overall_loss}
\end{equation}
where $\lambda_{\rm mar}$, $\lambda_{\rm sur}$, and $\lambda_{\rm con}$ are loss weights.
In practice, $\mathcal{L}_{\rm sur}$ and $\mathcal{L}_{\rm con}$ are enabled after the first 100 epochs to avoid unstable auxiliary supervision at the early stage of training.


\begin{figure}[!t]
    \centering
    \begin{subfigure}[t]{0.32\textwidth}
        \centering
        \includegraphics[width=\textwidth]{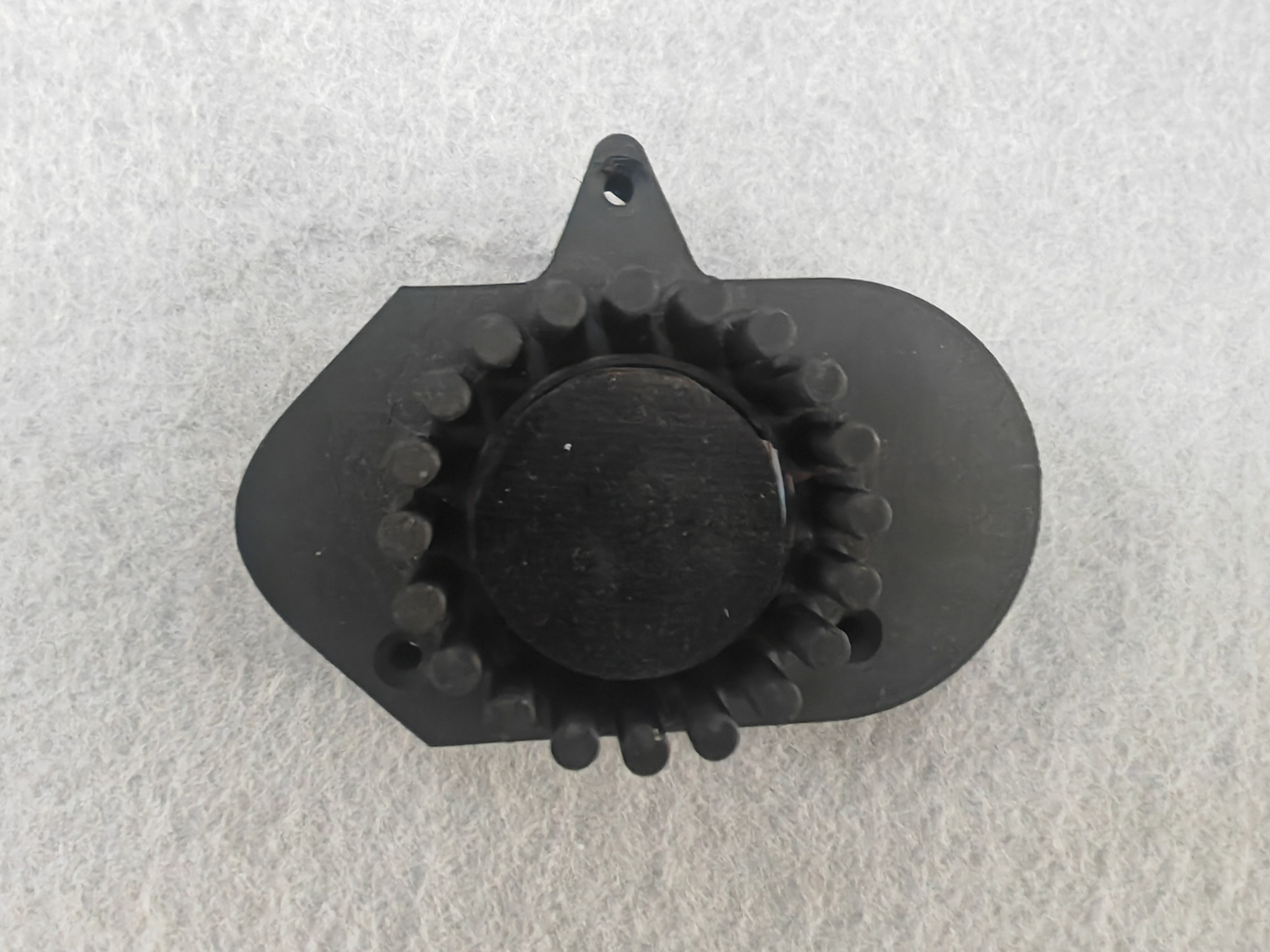}
        \caption{Sensor with alignment pins}
        \label{fig:sensor}
    \end{subfigure}
    \hfill
    \begin{subfigure}[t]{0.32\textwidth}
        \centering
        \includegraphics[width=\textwidth]{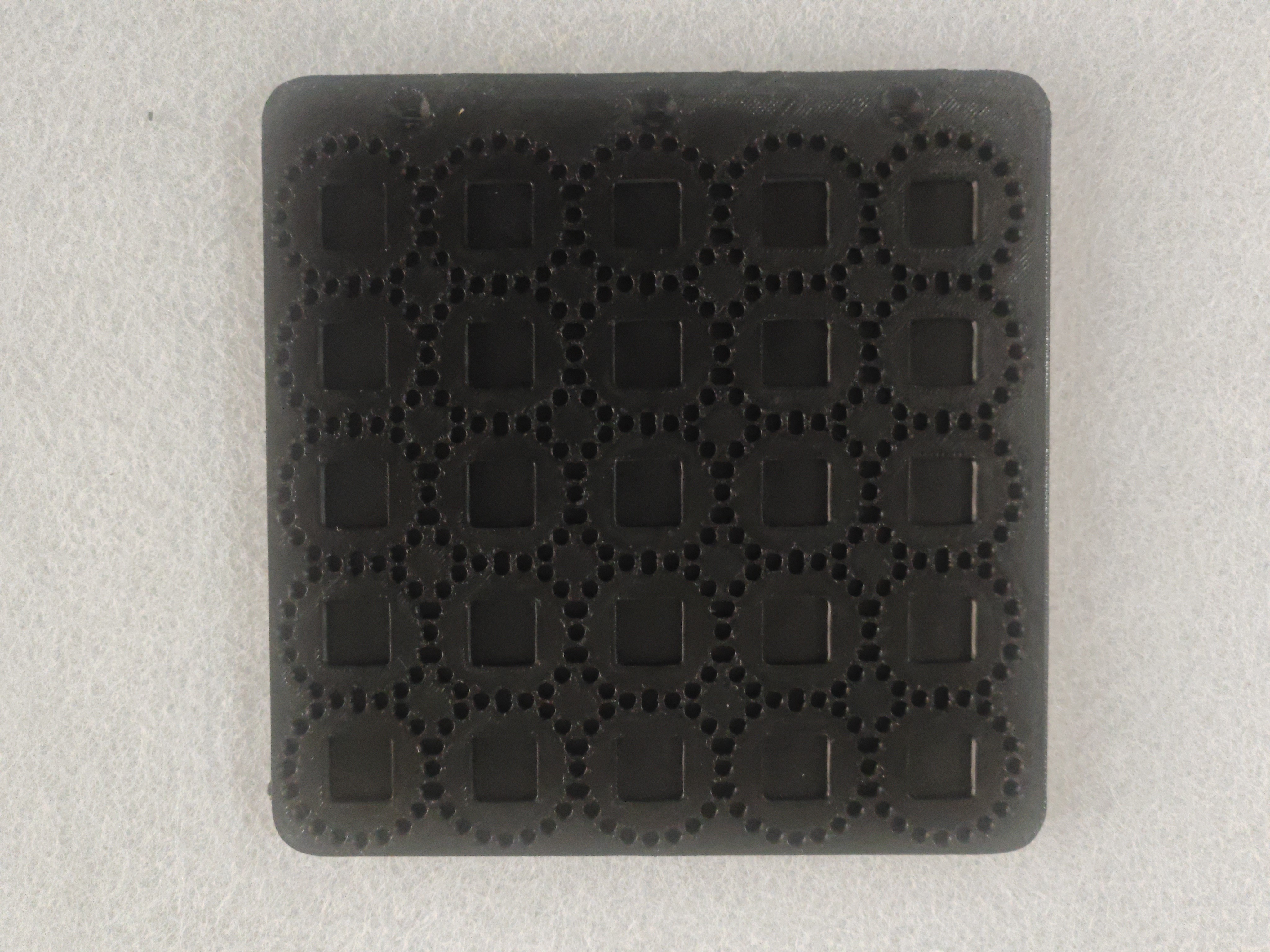}
        \caption{Collection pad}
        \label{fig:pad}
    \end{subfigure}
    \hfill
    \begin{subfigure}[t]{0.32\textwidth}
        \centering
        \includegraphics[width=\textwidth]{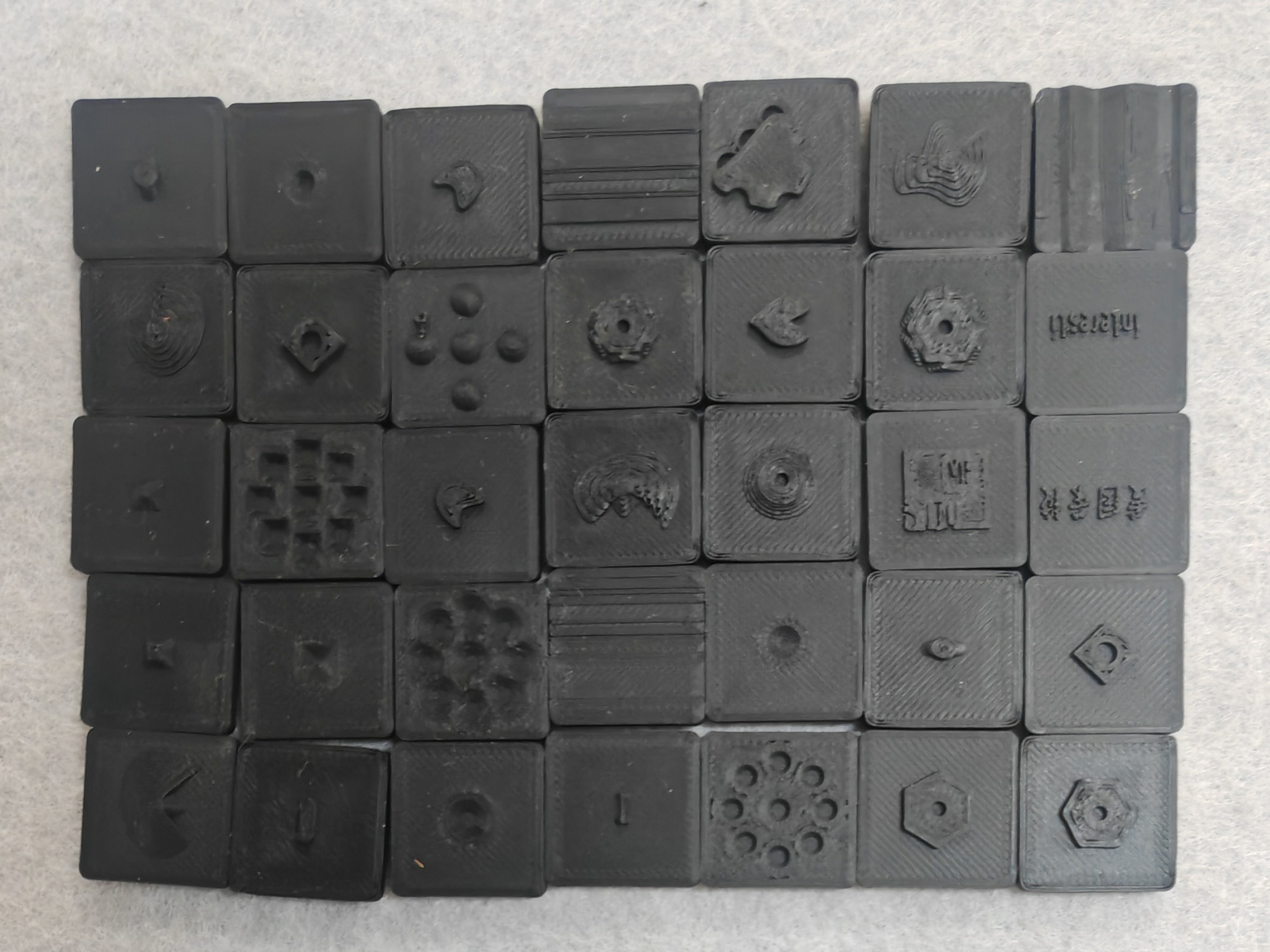}
        \caption{Texture unit samples}
        \label{fig:units}
    \end{subfigure}
    \caption{3D printed hardware for dataset collection.}
    \label{fig:supp_hardware}
\end{figure}

\subsection{Real-World Paired Dataset Collection}
\label{subsec: dataset}


Effectively training DeTac requires paired supervision for both shading restoration and marker extraction.
Existing synthetic datasets \citep{markerless24} naively superimpose markers onto marker-free images, which ignores real marker-induced indentation artifacts and introduces a synthetic-to-real gap. 
To bridge this gap, we design a hardware-assisted pipeline to collect a real-world, well-aligned paired dataset.
Specifically, we collect tactile data by repeatedly pressing the sensor against a diverse set of objects at multiple predetermined contact poses. 
Let $\M{I}_{\rm mix}^{o,k}$ denote the mixed tactile image acquired from object $o$ at the $k$-th pose using the marker-based sensor, with corresponding marker-free shading image $\M{I}_{\rm sha}^{o,k}$ and marker image $\M{I}_{\rm mar}^{o,k}$. 
The resulting dataset is formulated as
\begin{equation}
\mathcal{D}
=
\left\{
\left(
\M{I}_{\rm mix}^{o,k},
\M{I}_{\rm sha}^{o,k},
\M{I}_{\rm mar}^{o,k}
\right)
\;\middle|\;
o\in\mathcal{O},\;
k\in\{1,\dots,K\}
\right\},
\label{eq:dataset_definition}
\end{equation}
where $\mathcal{O}$ denotes the object set and $K$ denotes the number of repeatable contact poses per object. 


\mypara{Hardware-Assisted Paired Data Acquisition.}
To obtain highly aligned shading and mixed tactile images, we divide data acquisition into two stages. 
First, we collect marker-free shading images $\M{I}_{\rm sha}$ using a pristine sensor elastomer. 
We then fabricate markers directly on this elastomer surface to reduce fabrication discrepancy and acquire corresponding mixed images $\M{I}_{\rm mix}$.
Nevertheless, naively repeating the pressing process can still introduce pose drift.
To ensure consistent contact conditions across the two acquisition stages, we design a mechanical alignment structure, as shown in \cref{fig:hardware}.
Assisted by this structure, we follow the same pose sequence before and after marker fabrication, enabling high-quality paired data acquisition of real physical contacts.
We provide visualization of our 3D printed hardware for data collection, including the MC-Tac~\cite{mc_tac} sensor with alignment pins (\cref{fig:sensor}), the collection pad (\cref{fig:pad}), and a subset of texture units (\cref{fig:units}).


\mypara{Pseudo GT Marker Image Generation.}
After image acquisition, we generate pseudo GT marker images $\M{I}_{\rm mar}$ to supervise the marker extraction branch. 
To balance annotation efficiency and pixel-level accuracy, we use a prompt-based segmentation pipeline. 
We first manually annotate the center point of each marker in $\M{I}_{\rm mix}$. 
These point prompts, together with the original visual features, are then fed into SAM \citep{sam2} to segment the marker and produce $\M{I}_{\rm mar}$. 
Compared with fully manual pixel-wise annotation, this strategy provides reliable spatial priors with substantially lower annotation cost.

\mypara{Marker-Aware Augmentation for Cross-Sensor Generalization.}
To reduce overfitting to the fixed hardware specifications of our custom sensor and improve generalization across different marker-based tactile sensors, we introduce marker-aware spatial augmentations. 
Different sensor designs may exhibit different marker scales, densities, and boundary ambiguities due to variations in elastomer thickness and internal illumination. 
To simulate these variations, we apply aligned transformations to the triplet $(\M{I}_{\rm mix},\M{I}_{\rm sha},\M{I}_{\rm mar})$. 
Specifically, we use cropping followed by resizing to mimic changes in marker scale, and apply blurring guided by $\M{I}_{\rm mar}$ to simulate different levels of defocus. 
This strategy improves the cross-sensor robustness of DeTac.

\vspace{-2pt}
\section{Experiments}
\label{sec:experiments}

In this section, we conduct comprehensive experiments to evaluate DeTac. We first detail our experimental setup in \cref{subsec:setup}, followed by comparisons with baselines in \cref{subsec:comparison} and ablation studies in \cref{subsec:ablation}. Furthermore, we demonstrate how DeTac's decoupled outputs benefit downstream tasks in \cref{subsec:downstream}.
The appendix provides more results.

\subsection{Experimental Setup}
\label{subsec:setup}
\vspace{-1pt}

\mypara{Implementation Details.}
The whole pipeline is trained on a single NVIDIA RTX 4090 GPU for 300 epochs with a batch size of 2, which takes approximately 14 hours.
The shading restoration and marker extraction branches are optimized with learning rates of $10^{-5}$ and $10^{-4}$ respectively.
For the loss part, we use $\lambda_{\mathrm{mar}}=0.5$, $\lambda_{\mathrm{sur}}=0.05$, and $\lambda_{\mathrm{con}}=0.05$.
\\\mypara{Baselines.}
For shading restoration, we compare DeTac against: the fast-marching-based TELEA~\cite{telea}; two masked inpainting models, LaMa~\cite{lama} and the marker removal model TacDiff~\cite{markerless24}, both trained on synthetic data generated from our dataset following the latter's methodology; Stable Diffusion (SD) 1.5~\cite{stable_diffusion} fine-tuned on our data for full-image prediction; and three deterministic supervised baselines trained on the same paired dataset: a U-Net regression model, Pix2Pix, and a multi-task dual-head network.
Our TacDiff reproduction strictly follows its original training protocol, using synthetically superimposed markers.
We further evaluate enlarged-mask inpainting as an alternative strategy by dilating the marker masks to cover the surrounding indentation artifacts and applying them to both TacDiff and LaMa.
For marker extraction, we evaluate DeTac against UVTac~\cite{UVtac22} and the segmentation model SAM 3.1~\cite{sam3}. 
\begin{table}[!t]
    \centering
    \caption{Quantitative results of shading restoration. The best results are highlighted in \textbf{bold}.}
    \label{tab:image_metrics}
    \renewcommand{\arraystretch}{1.2}
    \setlength{\tabcolsep}{4pt} 
    
    \resizebox{\linewidth}{!}{
    \begin{tabular}{l cccc @{\hspace{1.5em}} cccc} 
        \toprule
        \multirow{2.35}{*}{Method} & \multicolumn{4}{c}{Unseen pose} & \multicolumn{4}{c}{Unseen object} \\
        \cmidrule(lr){2-5} \cmidrule(r){6-9}
        & FID $\downarrow$ & KID $\times 10^{-2}$ $\downarrow$ & PSNR $\uparrow$ & SSIM $\uparrow$ 
        & FID $\downarrow$ & KID $\times 10^{-2}$ $\downarrow$ & PSNR $\uparrow$ & SSIM $\uparrow$ \\
        \midrule
        
        TELEA \citep{telea}    & 237.95 & 18.327  & 31.33 & 0.870 & 232.66 & 17.592  & 31.29 & 0.868 \\
        TacDiff \citep{markerless24}           & 158.94 & 11.447  & 31.52 & 0.866 & 155.63 & 10.654  & 31.40 & 0.863 \\
        LaMa \citep{lama}   & 358.16 & 37.535  & 29.09 & 0.804 & 360.27 & 37.117  & 29.16 & 0.803 \\
        SD 1.5 \citep{stable_diffusion}  & 121.70 & 7.401   & 25.44 & 0.861 & 114.29 & 6.703   & 24.16 & 0.826 \\
        U-Net   & 108.35 & 7.489   & 38.56 & 0.950 & 111.28 & 7.750   & 39.03 & 0.942 \\
        Pix2Pix & 112.53  & 7.472 & 38.53 & 0.951 & 109.66  & 7.065 & 38.98 & 0.948 \\
        dual-head network & 104.27  & 6.923 & 37.99 & 0.949 & 105.18  & 6.794 & 38.37 & 0.947 \\
        TacDiff (enlarged mask) & 172.01 & 11.898 & 30.93   & 0.852  & 180.96 & 12.384   & 30.84 & 0.847 \\
        LaMa (enlarged mask) & 330.13  & 32.754 & 30.23 & 0.840 & 330.57  & 32.351 & 30.27 & 0.839 \\
        DeTac (w/o $L_{sur}$) & 25.78  & 0.733 & 34.64 & 0.946 & 22.59  & 0.698 & 35.05 & 0.945 \\
        DeTac (w/o $L_{con}$) & 22.65  & 0.689 & 37.77 & 0.949 & 19.44  & 0.634 & 38.28 & 0.948 \\
        DeTac (Full)       & \textbf{20.44} & \textbf{0.683} & \textbf{39.45} & \textbf{0.953} & \textbf{17.06} & \textbf{0.573} & \textbf{39.07} & \textbf{0.950} \\
        \bottomrule
    \end{tabular}
    }
    
    \vspace{-5pt}
\end{table}
\begin{table}[!t]
    \centering
    \caption{Quantitative evaluation of marker extraction. The best result is highlighted in \textbf{bold}.}
    \label{tab:mask_metrics_transposed}
    \setlength{\tabcolsep}{8pt} 
    \footnotesize
    \begin{tabular}{l ccccc}
        \toprule
        & UVTac \citep{UVtac22} & SAM 3.1 \citep{sam3} & DeTac (w/o $\mathcal{L}_{\rm sur}$) & DeTac (w/o $\mathcal{L}_{\rm con}$) & DeTac (Full) \\
        \cmidrule(lr){1-6}
        MCLE $\downarrow$ & 0.21 & 0.41 & 0.00839 & 0.00832 & \textbf{0.00762} \\
        \bottomrule
    \end{tabular}
    \vspace{3pt}
\end{table}
\begin{figure}[!t]
    \centering
    \includegraphics[width=\linewidth]{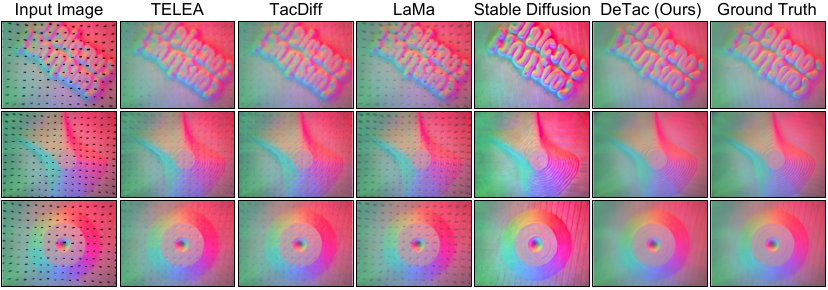}
    \caption{Qualitative comparison of shading restoration results.}
    \label{fig:visual_results}
\end{figure}
\\\mypara{Metrics.}
To evaluate the quality of the restored shading images, we adopt Fréchet Inception Distance (FID)~\citep{FID} and Kernel Inception Distance (KID)~\citep{KID} to measure distributional similarity with the GT shading images in a deep feature space, as well as Peak Signal-to-Noise Ratio (PSNR) and Structural Similarity Index Measure (SSIM)~\citep{SSIM} for pixel-level fidelity and structural similarity.
For marker extraction, we introduce the Marker Centroid Localization Error (MCLE).
For each marker, the predicted centroid $\hat{C}_i$ is computed as the average positions weighted by $\hat{\M{I}}_{\rm mar}$ within a circular region around the marker.
MCLE is then defined as the $L_2$ distance in \unit{\mm} between $\hat{C}_i$ and the GT centroid $C_i$ averaged over all markers, with lower values indicating more precise marker extraction.
\\\mypara{Dataset.} 
We set the number of poses per object to $K=20$ and the number of objects to $|\mathcal{O}|=54$, leading to 1080 pairs of tactile samples in total. 
We reserve 5 objects as ``Unseen objects'', and 1 pose from each of the other 49 objects as ``Unseen poses'' for testing. The rest are used for training.

\subsection{Evaluation Results}
\label{subsec:comparison}

\begin{figure}[!t]
    \centering
    \includegraphics[width=\linewidth]{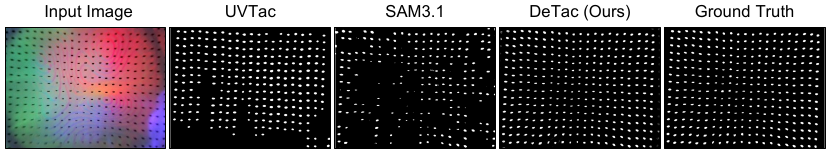}
    \vspace{-15pt}
    
    \caption{Qualitative comparison of marker extraction results.}
    \label{fig:comparison_mask}
    \vspace{5pt}
\end{figure}

\begin{figure*}[t!]
    \centering
    \begin{minipage}[t]{0.66\textwidth}
        \centering
        \begin{minipage}[c][2cm][c]{\textwidth}
            \centering
            \vspace{2pt}
            \includegraphics[width=\textwidth]{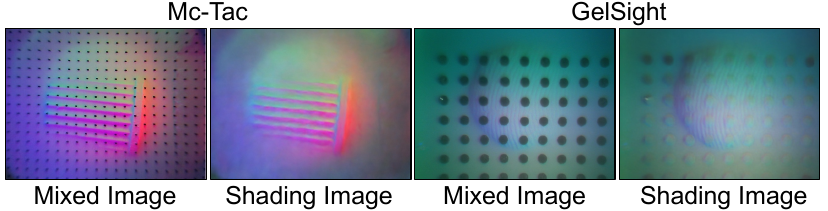}
        \end{minipage}
        \vspace{4pt}
        \caption{Generalization results of DeTac across different internal illuminations and sensor types.}
        \label{fig:generalization}
    \end{minipage}
    \hfill
    \begin{minipage}[t]{0.33\textwidth}
        \centering
        \begin{minipage}[c][2cm][c]{\textwidth}
            \centering
            \includegraphics[width=\textwidth]{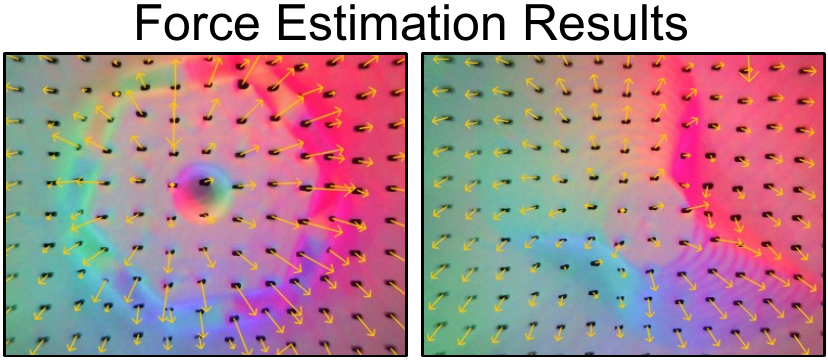}
    \vspace{-4pt}
            
        \end{minipage}
        \vspace{4pt}
        \caption{Force estimation results using extracted markers.}
        \label{fig:downstream}
    \end{minipage}
\end{figure*}

\mypara{Shading Restoration.} As shown in \cref{tab:image_metrics}, TELEA~\citep{telea}, TacDiff~\citep{markerless24}, and LaMa~\citep{lama} exhibit poor performance across all metrics. Stable Diffusion~\citep{stable_diffusion} achieves lower FID and KID but suffers from degraded PSNR and SSIM. The deterministic supervised baselines perform worse in perceptual and distributional similarity, indicating that DeTac better restores realistic tactile appearance and fine details. Enlarged-mask TacDiff and LaMa still achieve lower performance than DeTac. DeTac achieves the best performance across all metrics. Qualitatively, as shown in \cref{fig:visual_results}, TELEA~\citep{telea}, TacDiff~\citep{markerless24}, and LaMa~\citep{lama} leave visible residual marker traces, while Stable Diffusion~\citep{stable_diffusion} introduces unnatural stripe artifacts that distort the shading geometry. Our method produces the most natural results, closest to the GT.
\\\mypara{Marker Extraction.}
As shown in \cref{tab:mask_metrics_transposed}, UVTac~\citep{UVtac22} and SAM 3.1~\citep{sam3} exhibit significantly higher error compared to DeTac. 
\cref{fig:comparison_mask} also shows that UVTac~\citep{UVtac22} and SAM 3.1~\citep{sam3} produce coarse and incomplete masks, whereas DeTac successfully extracts all marker points with precise boundaries.
\\\mypara{Cross-sensor Generalization.}
We evaluate generalization of DeTac on 1) the same type of sensor with different internal illumination, and 2) a different type of sensor \citep{gelsight}.
As shown in \cref{fig:generalization}, DeTac restores high-quality shading in both settings, demonstrating cross-sensor generalization.

\begin{figure}[!t]
    \centering
    \includegraphics[width=\linewidth]{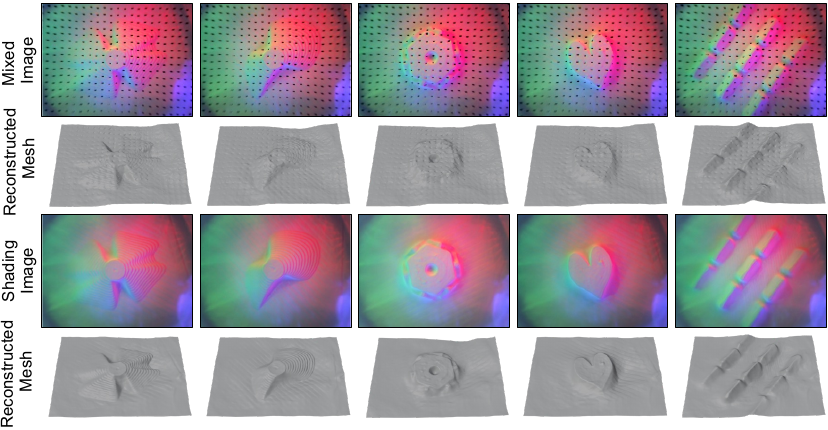}
    \vspace{-17pt}
    
    \caption{Surface texture reconstruction results.}
    \label{fig:recon}
\end{figure}

\vspace{-1pt}
\subsection{Ablation Study}
\label{subsec:ablation}

We ablate each loss term in \cref{subsec:optimization} by removing them individually.
As reported in \cref{tab:image_metrics} and \cref{tab:mask_metrics_transposed}, removing $\mathcal{L}_{\mathrm{sur}}$ causes the most severe degradation in shading quality on both novel poses and novel objects, confirming that explicit supervision on marker-neighboring regions is critical. Removing $\mathcal{L}_{\mathrm{con}}$ also degrades shading and slightly increases marker extraction error, indicating that the change map provides indispensable guidance for marker extraction. The full model achieves the best performance across all metrics, demonstrating that the two loss terms work synergistically.

\subsection{Downstream Tasks}
\label{subsec:downstream}

We demonstrate how DeTac's decoupled outputs benefit downstream tasks.

\mypara{Force Estimation.} Results in \cref{fig:downstream} demonstrate that our marker extraction enables precise displacement tracking for force estimation.
We further quantitatively evaluate force estimation. The tactile sensor is mounted on a robot end-effector equipped with a six-axis force sensor. The measured forces are transformed from the force-sensor coordinate frame into the tactile-sensor coordinate frame and used as ground truth. We separately train a force-estimation network for each method. The original mixed images and marker representations extracted by UVTac, SAM 3.1, and DeTac are fed into their respective networks and evaluated on the test set. As shown in \cref{tab:mask_metrics}, DeTac achieves the lowest mean force MAE, enabling more accurate force estimation than other methods.

\mypara{Surface Reconstruction.} As shown in \cref{fig:recon}, photometric stereo reconstruction~\citep{gs_sdk} using our restored shading image yields significantly more accurate surface meshes compared to using the mixed tactile image, where residual marker artifacts introduce obvious geometric distortions.
We further quantitatively evaluate surface reconstruction. The corresponding contact region extracted from the object’s 3D model is used as the ground truth for evaluating the reconstructed surface. Each reconstructed mesh is aligned with the ground truth before computing the Chamfer distance and RMSE. As shown in \cref{tab:shape_metrics}, reconstruction using DeTac shading images achieves lower reconstruction metrics than using mixed images, demonstrating improved surface reconstruction accuracy.

\begin{table}[t]
\centering
\begin{minipage}[t]{0.59\linewidth}
\centering
\caption{Force estimation results.}
\label{tab:mask_metrics}
\setlength{\tabcolsep}{3pt}
\resizebox{\linewidth}{!}{%
\begin{tabular}{@{}lrrrr@{}}
\toprule
 & Mixed & UVTac & SAM 3.1 & DeTac \\
\midrule
Mean force MAE (N) $\downarrow$ & 0.202 & 0.119 & 0.482 & \textbf{0.089} \\
\bottomrule
\end{tabular}%
}
\end{minipage}\hfill
\begin{minipage}[t]{0.40\linewidth}
\centering
\caption{Surface reconstruction results.}
\label{tab:shape_metrics}
\setlength{\tabcolsep}{3pt}
\resizebox{\linewidth}{!}{%
\begin{tabular}{@{}lrr@{}}
\toprule
 & Mixed & Shading \\
\midrule
Chamfer Dist. (mm) $\downarrow$ & 0.165 & \textbf{0.113} \\
RMSE (mm) $\downarrow$ & 0.111 & \textbf{0.071} \\
\bottomrule
\end{tabular}%
}
\end{minipage}
\vspace{5pt}
\end{table}

\section{Conclusion}
\label{sec:conclusion}
We presented DeTac, a dual-branch full-image generation framework that concurrently recovers shading and marker components from a mixed tactile observation.
A cross‑branch synergistic optimization mechanism couples both branches for mutual supervision, while a hardware‑assisted alignment structure provides paired real‑world data.
Experiments confirm DeTac's state‑of‑the‑art decoupling accuracy, strong downstream performance, and effective cross‑sensor generalization.

\noindent\textbf{Limitations.}
First, the shading branch requires hundreds of sequential sampling steps at inference, limiting real-time deployment on latency-sensitive manipulation tasks; promising remedies include accelerated samplers or distillation-based compression.
Second, our data acquisition pipeline is irreversible: markers are fabricated directly onto the elastomer and cannot be removed, so scaling the dataset requires manufacturing new sensors, which incurs additional hardware and labor overhead.
Third, the current data collection setup does not support controlled shear forces, limiting the training data distribution.

\par
\acknowledgments{This work is supported by National Natural Science Foundation of China (Grant No. 52505029), Science and Technology Commission of Shanghai Municipality (Grant Nos. 25ZR1401191 and 24511103400), and Beijing High Innovation Plan (Contract No. 202604841443).}

\clearpage
\bibliography{example}
\clearpage
\appendix
\setcounter{figure}{0}
\renewcommand\thefigure{A\arabic{figure}}
\setcounter{table}{0}
\renewcommand\thetable{A\arabic{table}}  
\setcounter{equation}{0}
\renewcommand\theequation{A\arabic{equation}}

\clearpage

\maketitlesupplementary
In this appendix, we provide additional implementation details in \cref{sec:supp_implement}, including a detailed description of the network architecture and training configurations. We further present a visualization of the paired dataset in \cref{sec:supp_dataset}, along with additional experiments on a real-world downstream fabric classification task in \cref{sec:supp_experiment}.

\section{Additional Implementation Details}
\label{sec:supp_implement}
\subsection{Network Architecture Specifications}
\mypara{Shading Restoration Branch.} The shading restoration network $D_{\theta}$ adopts a U-Net architecture with 22 residual blocks~\cite{resnet16}. It directly processes a 6-channel input --- formed by the pixel-space concatenation of the mixed image $\M{I}_{\rm mix}$ and the noisy target $\V{x}_t$ --- to output a 3-channel noise prediction. 
Specifically, the encoder features 4 downsampling stages with block output channels of (64, 128, 256, 256), where each stage contains 2 residual blocks. 
The bottleneck consists of 2 residual blocks. 
The decoder features 4 corresponding upsampling stages; to effectively fuse the concatenated skip connections, each decoder stage employs 3 residual blocks.
Every residual block utilizes 2 convolutional layers paired with Swish activations~\cite{swish18} and Group Normalization~\cite{groupnorm18}. Finally, a terminal convolutional layer projects the features to the target 3-channel noise space.

\mypara{Marker Extraction Branch.} The marker extraction network $M_{\phi}$ is constructed as a lightweight fully convolutional network~\cite{FCN} comprising three layers. 
The channel dimensions progress as $3 \rightarrow 32 \rightarrow 64 \rightarrow 1$, with each convolution using a kernel size of 3, a stride of 1, and a padding of 1. ReLU activations~\cite{ReLU10} follow the first two convolutions, while the final convolution outputs logits without an activation function. The soft marker image is obtained by applying a sigmoid function to these logits. During training, \texttt{BCEWithLogitsLoss} directly operates on the logits for numerical stability. 

\subsection{Additional Training Details}
For the denoising diffusion forward process, the total number of discrete training timesteps is set to $T = 1000$. 
The morphological dilation radius used to define the annular neighborhood $\M{R}_{\rm mar}$ for the surrounding indentation loss $\mathcal{L}_{\rm sur}$ is set to $r=2$ pixels. 
Optimization is performed using the AdamW optimizer~\cite{AdamW19} with $\beta_1 = 0.9$, $\beta_2 = 0.999$, $\epsilon = 10^{-8}$, and weight decay $= 0.01$.

\subsection{Data Collection Details}
\begin{figure}[htbp]
    \centering
    \includegraphics[width=\linewidth]{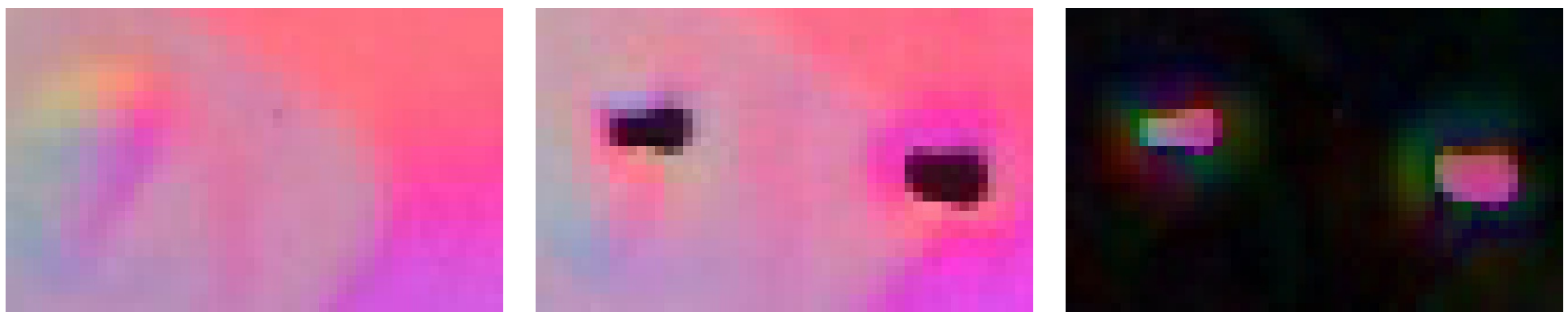}
    \caption{Pixel-wise errors caused by indentation artifacts.}
    \label{fig:indentation_error}
\end{figure}

During paired-data collection, 20 alignment pins fix pose and press depth, while a 200\,g weight maintains consistent press force. To evaluate the effect of press-force variation, images captured with 200\,g and 500\,g weights at identical depths yield a per-pixel MAE of 3.072 (0--255 scale), indicating a constant 200\,g force does not significantly limit dataset visual diversity. Markers are directly embedded in the same elastomer to minimize fabrication-induced pattern shifts.
In \cref{fig:indentation_error}, the left shading image, middle mixed image, and right image (their pixel-wise difference) reveal indentation artifacts around markers.

\section{Paired Dataset}
\label{sec:supp_dataset}

\begin{figure}[htbp]
    \centering
    \includegraphics[width=\textwidth]{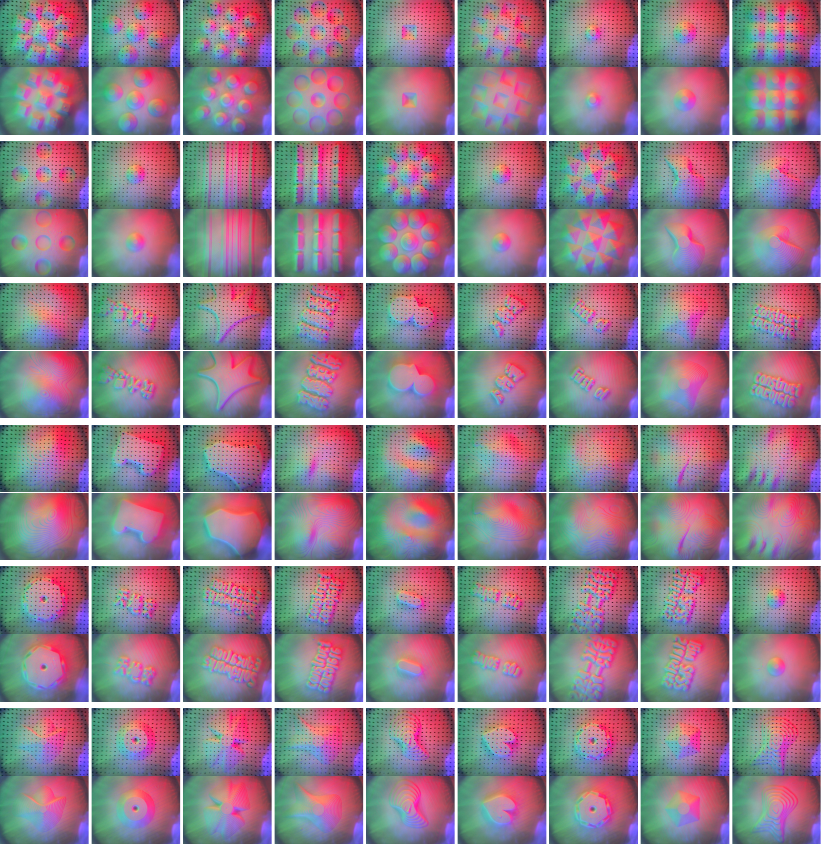}
    \caption{Visualization of paired data samples from each of the 54 texture units, arranged in 12 rows and 9 columns. Odd rows display the mixture image $\M{I}_{\rm mix}$ with markers, and even rows show the corresponding shading image $\M{I}_{\rm sha}$ directly below each mixture image.}
    \label{fig:paired_data}
\end{figure}

We further visualize in \cref{fig:paired_data} one paired data sample from each of the 54 texture units, resulting in 54 paired data examples in total.

\section{Additional Experiments}
\label{sec:supp_experiment}
To further verify that our model successfully decouples the signal, we conduct a real-world downstream fabric classification experiment to assess whether the separated shading image removes marker-induced interference while preserving sufficient physical information for practical tasks. We first train a ResNet-based~\cite{resnet16} texture classification network on data collected from a markerless tactile sensor. The sensor is mounted on a RealMan RM-75 robotic arm~\cite{RealmanRM75}, which presses against a fabric sample to perceive its surface texture and classify it among several fabric categories.

\begin{table}[htbp]
    \centering
    \caption{Fabric texture classification success rates across three fabric types.}
    \label{tab:fabric_classification}
    \begin{tabular}{cccccc}
        \toprule
        \multirow{2}{*}{Input Type} & \multicolumn{3}{c}{Success Rate} & \multirow{2}{*}{Overall} \\
        \cmidrule(lr){2-4}
        & Fabric A & Fabric B & Fabric C & \\
        \midrule
        Raw mixed image                     & 16.7\% & 43.3\% & 90.0\% & 50.0\% \\
        Our marker-removed shading image    & \textbf{83.3}\% & \textbf{73.3}\% & \textbf{96.7}\% & \textbf{84.4}\% \\
        \bottomrule
    \end{tabular}
\end{table}

We then replace the sensor with a marker-based sensor that is largely aligned with the markerless one, differing only in slight hardware variations. Using the same pre-trained classification network, we compare two evaluation settings: (1) directly feeding the raw mixed image into the classifier, and (2) feeding the marker-removed shading image predicted by DeTac into the classifier. For each of the three fabric types, we conduct 30 trials. The success rates are reported in \cref{tab:fabric_classification}. The results demonstrate that our method effectively eliminates marker interference, and the resulting shading images can be directly leveraged for downstream tasks. Although the robot arm carries a wrist-mounted RealSense D405C camera, its data is disregarded during experiment, our classification decisions are based exclusively on tactile signals.

\end{document}